%% file: main.tex
\documentclass[letterpaper, 10 pt, conference]{ieeeconf}  

\IEEEoverridecommandlockouts                              

\usepackage{graphicx} 
\usepackage{epsfig} 
\usepackage{times} 
\usepackage{amsmath} 
\usepackage{amssymb}  
\usepackage[T1]{fontenc}
\usepackage{color}
\usepackage{booktabs}
\usepackage[caption=false]{subfig}

\title{\LARGE \bf
Training Object Detectors With Noisy Data 
}

\author{Simon Chadwick and Paul Newman
\thanks{Authors are from the Oxford Robotics Institute, Dept. Engineering Science, University of Oxford, UK.
\texttt{\{simonc,pnewman\}@robots.ox.ac.uk}}}

\begin{document}

\maketitle
\thispagestyle{empty}
\pagestyle{empty}

\begin{abstract}

The availability of a large quantity of labelled training data is crucial for the training of modern object detectors. Hand labelling training data is time consuming and expensive while automatic labelling methods inevitably add unwanted noise to the labels. We examine the effect of different types of label noise on the performance of an object detector. We then show how co-teaching, a method developed for handling noisy labels and previously demonstrated on a classification problem, can be improved to mitigate the effects of label noise in an object detection setting. We illustrate our results using simulated noise on the KITTI dataset and on a vehicle detection task using automatically labelled data.

\end{abstract}

\section{Introduction}

Modern object detectors based on convolutional neural networks (CNNs) such as Faster RCNN \cite{ren2015faster}, RetinaNet \cite{lin2017focal} and SSD \cite{liu2016ssd} require a large quantity of labelled training data to achieve high performance. Manually labelling training data is expensive and time consuming, so assembling a dataset for a new task can be a significant hurdle to implementation. One approach that is often taken is to pre-train with one of a number of freely available established datasets (such as ImageNet \cite{russakovsky2015imagenet}) before fine-tuning using a small manually labelled dataset specific to the new task. However, this may not be possible if other sensor modalities are to be used or labelling a small fine-tuning dataset is not practical.

An alternative approach to the labelling problem is to use an automated labelling method. By using information not available at test time (such as expensive sensor modalities \cite{bruls2018mark}) automated (or semi-automated) labelling can provide labels of sufficient quality for training. However, automatically generated labels inevitably include noise which reduces the performance of models trained using the data.
   
A number of methods have been proposed that aim to reduce the impact of noisy labels when training CNNs (for example \cite{ren2018learning}). However, the focus of these works is often on classification tasks, with their efficacy being evaluated using small scale datasets such as MNIST and CIFAR. Some methods attempt to estimate the noise transition matrix which makes them hard to use in settings with other forms of noise (such as the size of a bounding box label in an object detection task).

The recently proposed co-teaching algorithm \cite{han2018co} is designed for training CNNs from noisy data. It has been shown to be effective in the presence of large quantities of label noise in a classification setting but the technique is agnostic to the specifics of the problem. This makes it suitable for deployment on object detection tasks and we show how the specifics of the task can be leveraged to modify the algorithm to further improve performance.

In this work we make the following contributions:
\begin{itemize}
    \item We examine the effect of different types of label noise on a typical CNN object detector.
    \item We document the change in performance as a result of applying co-teaching and modify the framework for better results.
    \item We apply these techniques to a real-world object detection problem for which a dataset including multiple sensor modalities has been collected.
\end{itemize}

\begin{figure}
\centering
\includegraphics[width=\columnwidth]{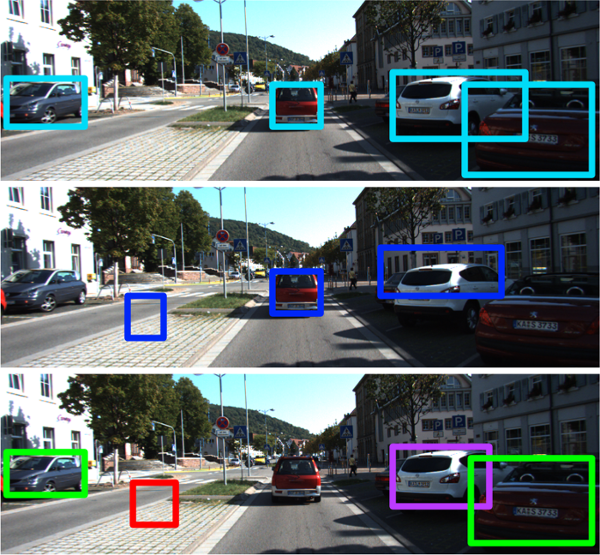} 
\caption{Our method adapts the co-teaching framework \cite{han2018co} for training object detectors from noisy data. The method significantly reduces the number of erroneous updates from losses due to noisy labels. A toy example is shown in the image. \textit{Top} shows the correct ground truth boxes, \textit{middle} shows labels that might result from a noisy labelling process. \textit{Bottom} shows the losses that our modified co-teaching method attempts to identify, preventing the corresponding weight updates from inhibiting the learning process. Green losses are those due to the absence of an object. Red updates are those that would reinforce the presence of an object due to an extra ground truth box. Purple losses are those that indicate a mismatch between the predicted box size and the ground truth.}
\label{fig:missing_boxes}
\end{figure}

\section{Related Work}
Reducing the need to hand-label training data by taking advantage of large quantities of unlabelled, weakly labelled or automatically labelled data has been a topic of research for a long time. 

One option is to attempt to label the unlabelled data during training. The well known co-training algorithm \cite{blum1998combining} iteratively increases the pool of labelled data by labelling unlabelled examples using a pair of classifiers trained on different views of the data in the previous pool. Another method \cite{rosenberg2005semi} that, like co-training, pre-dates the deep learning era, iteratively adds unlabelled data to the labelled set by scoring detections based on a detector independent metric.

More recently, methods such as \cite{misra2015watch} and \cite{jin2018unsupervised} exploit temporal consistency in videos to automatically mine new examples using an existing weak detector.

\cite{rasmus2015semi} demonstrates how sharing weights between an auto-encoder-like ladder network on unlabelled data with the feature extractor used for labelled data can leverage the unlabelled data during training to increase performance.

An alternative approach that has been gaining popularity is the use of synthetic training data. The effectiveness of this approach is demonstrated in \cite{tremblay2018training} where the use of synthetic data achieves good results, however it also shows the value of a small quantity of real data for fine-tuning. 

If the data already has labels but those labels are noisy there are a number of approaches that have been proposed to reduce the influence of the noise on the training process. Bootstrapping \cite{reed2014training}, for example, incorporates the current state of the model into the loss term resulting in down-weighting of the influence of the label for examples for which the model is confident.

\cite{sukhbaatar2014training} introduces two methods for estimating the noise transition matrix in classification problems.  The S-model \cite{goldberger2017training}, also attempts to learn the noise transition matrix.

Countering the view that noise needs to be carefully modelled,  \cite{krause2016unreasonable} shows that a large quantity of noisy data can be effectively used to increase performance on a fine-grained classification task. From a similar perspective, \cite{mahajan2018exploring} attempts to find the limits of weakly supervised data by pre-training with 3.5 billion images for classification and object detection tasks.

\section{Co-teaching}

Co-teaching \cite{han2018co} is based on the memorisation characteristics of neural networks \cite{arpit2017closer}. This effect means that simple patterns are learnt before examples are memorised. Provided that fewer than half of the labels are noisy (i.e. the labels don't form their own more dominant pattern), those examples with noisy labels won't fit the prevailing pattern and so will be more difficult to learn. Consequently, after a period of training, harder (and therefore more likely to be noisy) examples will have higher loss values.

\begin{figure}
\vspace{1em}
\centering
\includegraphics[width=\columnwidth]{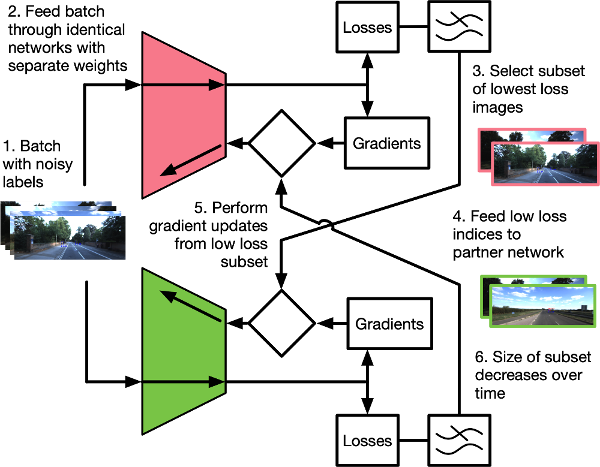} 
\caption{The standard co-teaching process proposed in \cite{han2018co}.}
\label{fig:co-teaching}
\vspace{-1em}
\end{figure}

To exploit this effect, co-teaching uses a pair of networks trained in parallel. The same batch is passed through both networks with the gradient updates of only those examples that are classified as low loss by the other network being applied. The proportion of examples in the batch that are excluded is set according to a schedule, increasing from zero at the start of training to a proportion corresponding to the estimated proportion of noisy labels. A schematic of the algorithm is shown in Fig. \ref{fig:co-teaching}.

The two key hyper-parameters that need choosing are the estimated proportion of noisy labels and a time constant, the number of epochs over which to reduce the proportion of selected examples --- if this constant is too large, the network will learn noisy examples before they can be effectively excluded, too small and the network has not had time to learn patterns before the loss values start being used to choose which examples to exclude.

\subsection{Types of label noise}

The original co-teaching work demonstrates the technique's efficacy in a classification setting with two types of label noise: symmetry noise, where a fixed proportion of labels are replaced by any of the alternative labels with equal probability and pair noise, where a fixed proportion of labels are consistently replaced with the next label. Another type of label noise that is not considered but which is of particular relevance when using automated labelling method, is systematic noise under which the labels of classes with similar appearance are swapped (for example 8 with 3 in the case of MNIST).

In an object detection setting each label comprises a bounding box and an associated class label. In addition to the types of noise already discussed that may affect the class label, there are additional types of label noise that may occur: spurious bounding boxes with random (or systematic) class labels may be present, bounding boxes may be incorrectly sized or positioned or may be missing entirely.
\subsection{Co-teaching for object detection}

\begin{figure*}[tpb]
    \centering
    \mbox{}\hfill
    \subfloat[Standard co-teaching\label{fig:standard-struct}]{\includegraphics[width=0.75\columnwidth]{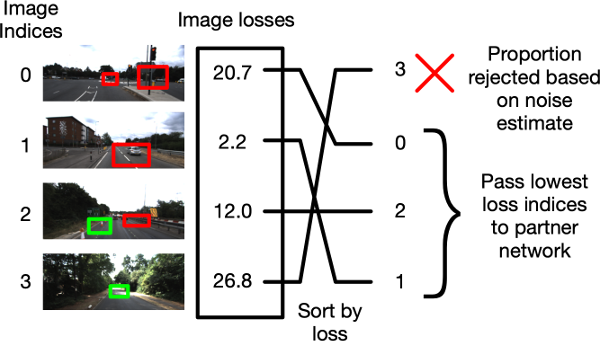}}\hfill
    \subfloat[Per-object co-teaching\label{fig:three-struct}]{\includegraphics[width=\columnwidth]{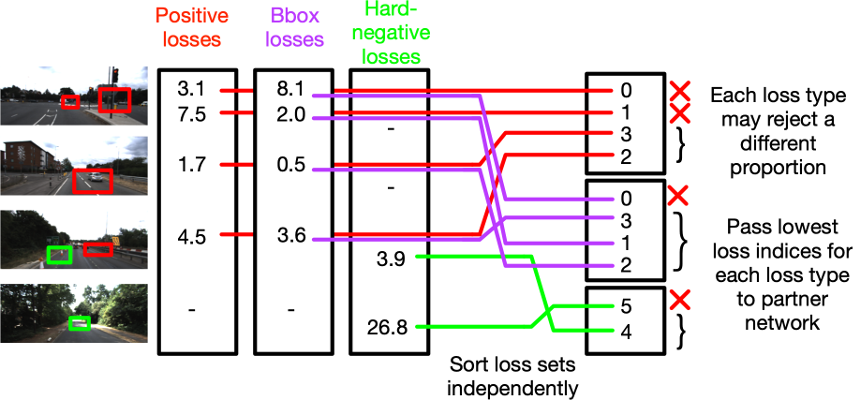}}
    \hfill
    \mbox{}
    \caption{A comparison between the loss selection structures of standard co-teaching and our proposed per-object formulation. In the standard method the losses of all objects in an image are collected together potentially mixing together clean and noisy labels. Our method allows each object in an image to be treated separately based on the noise affecting only its label.}
    \label{fig:loss-structure}
\end{figure*}

In the standard co-teaching formulation, gradient updates are applied for some images in a batch while the gradient updates from the highest loss images are excluded. This framework works effectively in a classification setting where the label applies to the whole image, so if the label is noisy it is reasonable to discard all updates related to that image. In an object detection setting, each image can contain multiple labels only some of which may be noisy. Consequently, by excluding the whole image, potentially useful gradient updates are being discarded.

\begin{figure}
\centering
\includegraphics[width=\columnwidth]{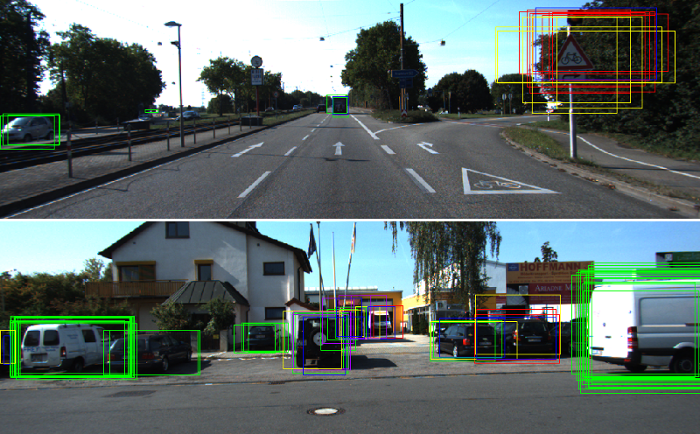} 
\caption{Examples of our per-object co-teaching process in action. In the images above, the labels are noisy (for example a number are missing) and are shown in blue. The other boxes are all losses from anchor boxes that have been removed by co-teaching as potentially erroneous. Green boxes are hard-negatives i.e. those that indicate the absence of an object, consequently any hard-negatives that actually appear on objects would have a detrimental effect on the learning process. Red boxes are those for which the predicted class is significantly different to the label (i.e. the positive cross-entropy is large). Purple boxes have large bounding box errors and yellow boxes are those that have both large positive cross-entropy and large bounding box errors.}
\label{fig:ranking_examples}
\end{figure}

We therefore introduce a modified formulation where gradient updates are selected on an object-by-object basis rather than an image-by-image basis. The simplest solution would be to rank total losses per object (by object, in this context we are referring to any anchor box that has been selected to contribute to the loss either by being matched to a label or selected as a negative). However, this straightforward modification is not well-posed in the context of training an SSD-style object detector and results in unstable training. This is because the SSD loss is formed of three parts --- cross-entropy losses for positives, cross-entropy losses for hard-negatives and bounding box losses for positives. In the naive object-by-object formulation, the two networks suffer no penalty for constantly predicting all negatives and entirely excluding all positive examples on the basis that their high losses are a result of noisy labels. The exclusion of positive losses is exacerbated by the SSD loss in which the number of sampled hard-negatives is three times that of the number of positives (subject to a hyperparameter). To overcome this we construct our per-object formulation by selecting subsets of low loss examples independently for each component of the loss. This ensures that there are always losses associated with positives and hard-negatives contributing to each update step. It also means that each of the loss components can have a different estimated noise fraction and therefore each can have a low loss subset that constitutes a different fraction of the available losses. For example, if the prevailing noise is predominantly in bounding box size and position, the fraction of bounding box losses left out can be larger than the fractions for positive or hard-negative cross entropy. An illustration of the difference between the standard formulation and our proposed method is shown in Fig. \ref{fig:loss-structure}. To gain more insight into the behaviour of our method the losses that are being removed can be plotted, as shown in Fig. \ref{fig:ranking_examples}.

Our experimental results in section \ref{sec:results} demonstrate that this per-object formulation not only improves performance in the presence of label noise but also when compared to employing the standard per-image co-teaching formulation.  

\section{Automatic Label Generation}\label{sec:generation}

A motivating force behind this work is to enable the use of automatically labelled datasets. Being able to use automatically labelled data reduces the barrier to deployment of supervised learning systems in situations where existing hand-labelled datasets are not suitable. This might occur when operating in a domain that is not well served by existing datasets (such as at night) or when additional sensor modalities are required. Unfortunately, due to imperfections in labelling systems, automatically labelled datasets inevitably contain labelling errors hence the need for the method we are proposing. 

To validate our approach we recorded and automatically labelled a new vehicle object detection dataset (full details on this dataset are given in \cite{chadwick2019distant}). On a series of drives we recorded stereo wide-angle RGB images, monocular zoom lens images as well as radar target data from a cruise control radar.

\begin{figure}
\vspace{1em}
\centering
\includegraphics[width=\columnwidth]{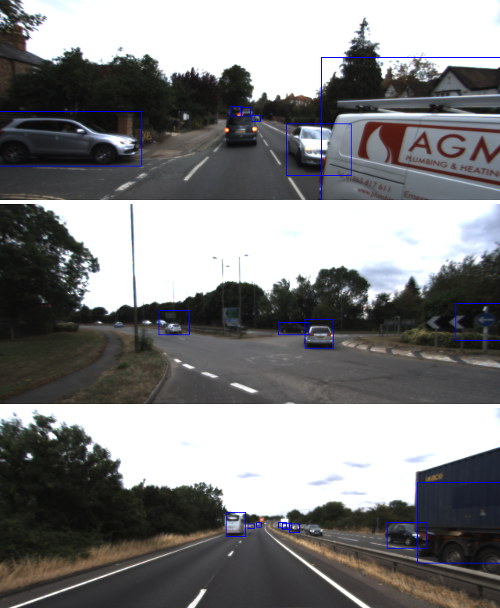} 
\caption{Some example data from the automatically labelled dataset illustrating the noise in the labelling process. In \textit{top} note the missing bounding box for the car immediately in front and in \textit{middle} the mis-labelled sign. \textit{Bottom} shows some missing and mis-sized bounding boxes as well as the range of vehicle sizes.}
\label{fig:example_labels}
\vspace{-1em}
\end{figure}

To perform the labelling we ran an existing object detector (based on a YOLO \cite{redmon2016you} architecture) over both the wide-angle and zoom lens images. We then combined the detections from the left wide-angle camera and the zoom lens camera by exploiting the fact that the cameras were physically located as close as possible to each other. This allowed us to transfer the long range detections from the zoom lens into the wide-angle images, improving the quality of the labels in the overlapping image region (which largely corresponds to the area of the image where most long range objects occur). We also use a visual odometry system \cite{churchillvo} to generate ego-motion estimates from the stereo pair which are used to subtract the ego-motion from the radar target range rates (radial velocities). Some example labelling from the dataset is shown in Fig. \ref{fig:example_labels}.

The resulting dataset contains 17553, 2508 and 5015 frames each for training, validation and testing respectively. Each frame consists of the left wide-angle image, the ego-motion adjusted radar data and the generated class labels and bounding boxes.

\section{Network setup}

To demonstrate the effect of noisy labels and co-teaching we conduct experiments using a standard object detector framework based on SSD that we train from scratch for each experiment. One of the downsides of co-teaching is that two networks are trained simultaneously meaning that twice as much memory is required compared to training a normal network\footnote{As an upside the process produces two fully trained networks. In this work we only make use of the results from the first network although they could potentially be combined.}. As a result we use a relatively small ResNet18 \cite{he2016identity} backbone network. For more details of the network configuration see \cite{chadwick2019distant}.

Standard image augmentations (flips, crops and colour shifts) and L2 weight decay of 1e-3 are used during training. We train using the ADAM optimiser with an initial learning rate of 1e-4. The learning rate is reduced by a factor of 10 after 45k and 55k iterations.

\section{Experimental results}\label{sec:results}

We conduct experiments on two datasets, the KITTI 2D object detection dataset \cite{Geiger2012CVPR} and our own dataset described in Section \ref{sec:generation}.

The KITTI dataset includes images, labels and evaluation code for a 2D object detection benchmark. By adding controlled quantities of noise to this hand labelled dataset we can conduct a set of experiments to demonstrate the effect of label noise when training object detectors. As the labels for the test portion of the dataset are not available we conduct the experiments using splits of the training set with 5985, 374 and 1122 images for training, validation and testing respectively. For the KITTI experiments, all images are resized to 1024 x 320, training runs for 200 epochs using a co-teaching epoch constant of 50 and we evaluate using the KITTI evaluation code.

In all experiments we evaluate using average precision (AP). 

\subsection{Effect of label noise on KITTI}

We first examine the effect of different label noise types on object detector performance. We experiment with five types of noise:
\begin{enumerate}
\item\label{noise-whole} Whole image label noise. This is an approximation of the classification case such that with a certain probability per image, all the labels for ground truth objects in that image are corrupted using pair noise.
\item\label{noise-bbox} Per object bounding box noise. For each ground truth object labelled in an image, with a certain probability the bounding box of that label is shifted and scaled by amounts sampled from a normal distribution.
\item\label{noise-extra} Spurious additional bounding boxes. For each image, with a certain probability, an extra bounding box of random position and dimensions is added to the image with a label randomly selected from the list of classes.
\item\label{noise-missing} Missing bounding boxes. With a certain probability each ground truth bounding box may be removed from the set of labels.
\item\label{noise-combined} Combined bounding box noise. This applies noise types (\ref{noise-bbox}), (\ref{noise-extra}) and (\ref{noise-missing}) simultaneously.
\end{enumerate}
The results (Table \ref{tab:kitti_noise}) show that object detection performance is significantly affected by label noise. While it is hard to make comparisons across noise types given that the probability is applied in different ways (e.g. extra boxes are per-image while missing boxes are per-box), it is interesting to note that even with 50\% of the boxes missing performance only drops from 0.629 to 0.518 (this is in line with a result presented by \cite{wu2018soft}).

\begin{table}
    \vspace{1em}
    \centering
    \caption{Performance on KITTI dataset (Average precision on "Car, Moderate") with varying levels of noise of different types}
    \begin{tabular}{@{}l|ccc@{}}\toprule
     & \multicolumn{3}{c}{\textbf{Noise Probability}} \vspace{0.3em} \\
    \textbf{Noise Type} & \textbf{0.0} & \textbf{0.25} & \textbf{0.5}\\ \midrule
        None & 0.629 & - & -\\
        Whole image label noise (\ref{noise-whole}) & - & 0.506 & 0.385\\
        Per-object bounding box noise (\ref{noise-bbox}) & - & 0.577 & 0.502\\
        Additional boxes (\ref{noise-extra}) & - & 0.560 & 0.587\\
        Missing boxes (\ref{noise-missing}) & - & 0.593 & 0.518\\
        Combined noise (\ref{noise-combined}) & - & 0.457 & 0.317\\
    \bottomrule
    \end{tabular}
    \label{tab:kitti_noise}
\end{table}

\subsection{Co-teaching on KITTI}

We next compare the performance of the standard and per-object co-teaching algorithms using the KITTI dataset. We compare using two sorts of noise, whole image noise (\ref{noise-whole}) and combined noise (\ref{noise-combined}) as this is a reasonable approximation of the sort of noise present in our own dataset (see Section \ref{sec:generation} and Fig. \ref{fig:example_labels}). 

\begin{figure}
\vspace{1em}
\centering
\includegraphics[width=0.8\columnwidth]{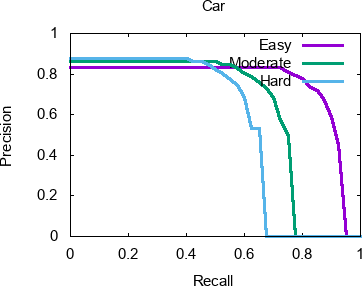} 
\caption{KITTI precision-recall curve for the car class using per-object co-teaching with labels corrupted by combined noise with a probability of 0.5.}
\label{fig:kitti-per-object}
\end{figure}

The results are shown in Table \ref{tab:kitti_coteach}. It can be seen that our proposed per-object co-teaching formulation improves performance over plain network training in every case and also outperforms standard co-teaching by a large margin in all but the simplest scenario. While it is initially interesting that one of the per-object co-teaching scores exceeds that of the no-noise case in Table \ref{tab:kitti_noise}, this result is not replicated in wider metrics. Fig. \ref{fig:kitti-per-object} shows the KITTI evaluation curve for cars with combined noise at a probability of 0.5 using per-object co-teaching.
\begin{table}\centering
    \caption{Comparison of standard and per-object co-teaching on the KITTI dataset (Average precision on "Car, Moderate")}
    \begin{tabular}{@{}l|c|ccc@{}}\toprule
    & & \multicolumn{3}{c}{\textbf{Co-teaching type}} \vspace{0.3em} \\
    \textbf{Noise type} & \textbf{Noise probability} & \textbf{None} & \textbf{Standard} & \textbf{Per-object}\\ \midrule
        Whole image (\ref{noise-whole}) & 0.25 & 0.506 & \textbf{0.617} & 0.613\\
        Whole image (\ref{noise-whole}) & 0.5 & 0.385 & 0.342 & \textbf{0.456}\\
        Combined (\ref{noise-combined}) & 0.25 & 0.457 & 0.527 & \textbf{0.666}\\
        Combined (\ref{noise-combined}) & 0.5 & 0.317 & 0.368 & \textbf{0.605}\\
    \bottomrule
    \end{tabular}
    \label{tab:kitti_coteach}
    \vspace{-1em}
\end{table}

\subsection{Effect of batch size}

\begin{figure}
\centering
\includegraphics[width=0.75\columnwidth]{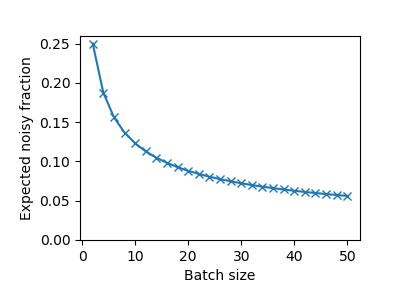} 
\caption{The effect of varying batch size on the expected fraction of noisy labels that remain in the subset of the batch that is selected by co-teaching.}
\label{fig:batch_size}
\end{figure}

One likely contributory factor in the outperformance of our per-object formulation over standard co-teaching for object detection relates to batch size. The basic operation in co-teaching is that a batch is sampled at random and then a certain fraction of the updates from the batch are discarded as noisy based on the ranked loss values. However, as the batch is randomly sampled and the fraction discarded is not adjusted based on the properties of the batch, there is a non-zero probability that some noisy updates remain in the pruned batch. If it is assumed that it is possible to perfectly identify noisy and non-noisy items, in the standard formulation the expected number of noisy images \(\mathbb{E}[n]\) that remain can be expressed as

\begin{equation}
    \mathbb{E}[n] = \displaystyle\sum_{k=0}^{N}\max(0, k - pk)\binom{N}{k}p^k(1-p)^{N-k} 
    \label{eq:batch_size}
\end{equation}

where the probability \(p = p_{i}\) the noise probability for each image and \(N = N_{i}\) the number of images in the batch. As shown in Fig. \ref{fig:batch_size}, by using a larger batch size the proportion of the remaining batch (\(\mathbb{E}[n] / N\)) that is expected to be noisy is decreased. If a per-object rather than per-image formulation is used then \(p = p_{O}\) the noise probability for each object and \(N = \phi N_{i}\) where \(\phi\) is the mean number of objects per image. As as result, assuming \(\phi > 1\), the effective batch size is larger and the expected fraction of the batch that is noisy is reduced.

\subsection{Co-teaching performance on automatically labelled data}

Having proven the effectiveness of our method when noise is artificially added, we employ our method on our automatically labelled dataset (see \ref{sec:generation}).

We employ the same network as for the KITTI tests with some minor adjustments. Due to the number of very small objects in the dataset we employ a technique described in \cite{muller2018detecting} to increase the number of default boxes used in each detection head. As the dataset is larger than KITTI we train for 70 epochs and the co-teaching epoch constant is set to 20.

As in \cite{chadwick2019distant} we conduct two forms of testing. We first conduct tests using the entire test split of the automatically labelled dataset. However, as the labels in the test set may also be noisy (and with potentially the same systematic biases as the training set) we also conduct tests using a small subset of the test set that has been hand-labelled.

For evaluation, as the KITTI evaluation ignores objects below a size threshold, again we follow \cite{chadwick2019distant}. This uses an IOU threshold of 0.5 and keeps all objects regardless of size. It also uses the more accurate AP integration method from \cite{pascal-voc-2012}. The maximum F1 score attainable at an operating point along the precision-recall curve is also calculated.

\begin{figure}
\vspace{1em}
\centering
\includegraphics[width=0.85\columnwidth]{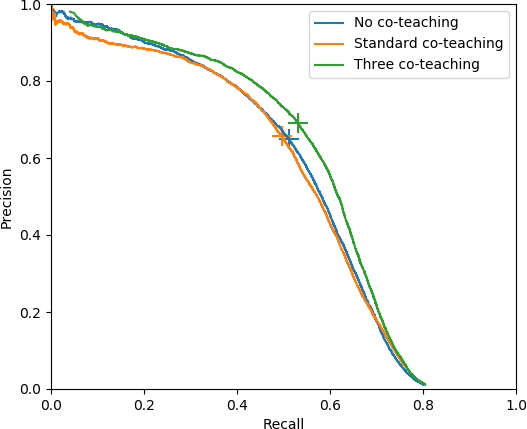}
\caption{Results on the automatically labelled dataset using the full automatically labelled (and therefore noisy) test set. The operating point corresponding to the maximum F1 score is also shown by a cross on each curve.}
\label{fig:auto_label_result}
\end{figure}

\begin{figure}
\centering
\includegraphics[width=0.95\columnwidth]{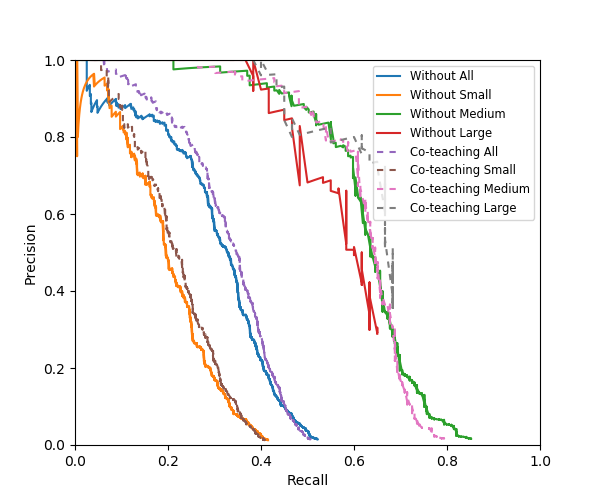} 
\caption{Comparison of the performance of a network trained without co-teaching (solid lines) and a network trained with our per-object co-teaching (dotted lines) on the hand-labelled subset of the test set from our dataset (see Section \ref{sec:generation}). Separate curves are shown for subsets of the objects categorised by object size.}
\label{fig:hand_ap_curves}
\end{figure}

The results of testing against both the full automatically labelled test set and the hand-labelled subset are shown in Table \ref{tab:hand_results}. It can be seen that our per-object co-teaching method consistently provides the best performance. The overall average precision curves for all three methods are shown in Fig. \ref{fig:auto_label_result}. It is worth noting that one of the reasons for the lower levels of performance shown by all methods on the hand-labelled subset is that this subset, by virtue of the labeller having access to the zoom lens images, contains a large number of very small objects. This effect can also be seen in Fig. \ref{fig:hand_ap_curves} where the "All" curve largely mirrors the shape of the "Small" curve.

\begin{table}
    \vspace{1em}
    \centering
    \caption{Performance when trained using automatically labelled data. Evaluated against both automatic and hand-labelled test sets}
    \begin{tabular}{@{}l|c|cc@{}}\toprule
    \textbf{Co-teaching type} & \textbf{Test set} & \textbf{AP} & \textbf{Max F1}\\ \midrule
        None & Automatic & 0.528  & 0.572\\
        Standard & Automatic & 0.517 & 0.566\\
        Per-object & Automatic & \textbf{0.555} & \textbf{0.601}\\
        \midrule
        None & Hand subset & 0.302 & 0.399\\
        Standard & Hand subset & 0.321 & 0.414\\
        Per-object & Hand subset & \textbf{0.327} & \textbf{0.417}\\
    \bottomrule
    \end{tabular}
    \label{tab:hand_results}
    \vspace{-1em}
\end{table}

\section{Conclusions and further work}

Learning from datasets with noisy labels is an important avenue of research as it makes the use of supervised machine learning models more viable in many applications. In this work we have shown how noise can affect the training of a CNN object detector and demonstrated how a modified version of co-teaching can be used to mitigate the effects of noisy labels. It is fair to say that this is not a perfect approach: the need to train two networks simultaneously places constraints on the types of models that can be easily used. In addition, there are a number of hyperparameters that have to be tuned to achieve any improvement in performance. At present the only methods for establishing which hyperparameters to use are manual inspection (for an initial estimate of noise parameters) and cross validation. For future work we plan to investigate ways of estimating the noise parameters and, by extension, estimates of the hyperparameters. One possible method for this is through inspection of the distribution of the various loss values during training.

Another interesting direction for future work would be to investigate the effect of dataset size. It is possible that the benefits of co-teaching diminish with larger datasets. 


\section*{Acknowledgements}
The authors would like to thank Martin Engelcke and colleagues for the object detector implementation used to generate the dataset. We also gratefully acknowledge the JADE-HPC facility for providing the GPUs used in this work. Paul Newman is funded by the EPSRC Programme Grant EP/M019918/1.


\bibliographystyle{IEEEtran}
\input{bibliography.bbl}


\end{document}